\documentclass[pdflatex,sn-basic,Numbered]{sn-jnl}

\usepackage{graphicx}%
\usepackage{multirow}%
\usepackage{amsmath,amssymb,amsfonts}%
\usepackage{mathrsfs}%
\usepackage[title]{appendix}%
\usepackage{xcolor}%
\usepackage{textcomp}%
\usepackage{manyfoot}%
\usepackage{booktabs}%
\usepackage{listings}%
\usepackage{placeins}%

\theoremstyle{thmstyleone}%
\theoremstyle{thmstyletwo}%
\theoremstyle{thmstylethree}%
\begin{document}

\title{A Step Harder, Not a Step Too Far: The Benefits and Limits of Curriculum Difficulty Progression in GRPO}

\author[1]{\fnm{Ronghan} \sur{Li}}

\author*[2]{\fnm{Fei} \sur{Chen}}\email{chenf@dlmu.edu.cn}

\author[1]{\fnm{Mingyu} \sur{Lu}}

\affil[1]{\orgdiv{College of Artificial Intelligence},
\orgname{Dalian Maritime University},
\orgaddress{\city{Dalian}, \postcode{116026},
\state{Liaoning}, \country{China}}}

\affil*[2]{\orgdiv{Information Science and Technology College},
\orgname{Dalian Maritime University},
\orgaddress{\city{Dalian}, \postcode{116026},
\state{Liaoning}, \country{China}}}
\abstract{
Curriculum learning typically organizes training samples from easy to hard. In GRPO, medium-difficulty samples are attractive for online selection because they are more likely to yield informative group-relative learning signals, and prioritizing them can improve training efficiency. However, it remains unclear whether persistently selecting such samples also leads to the strongest final model performance, or whether training can benefit from progressing toward harder samples. Meanwhile, sample difficulty changes as the policy evolves, so fixed difficulty estimates can become misaligned with the current policy. We therefore re-estimate sample difficulty under the current policy and, inspired by the easy-to-hard principle of curriculum learning, design and evaluate a staged curriculum that progressively shifts the active difficulty band toward harder samples. Experiments on GSM8K show that persistent medium-difficulty selection does not achieve the strongest observed performance. Across three random seeds, progressing to a moderately harder difficulty band improves test accuracy to 0.6591 ± 0.0077, compared with 0.6449 ± 0.0024 for the fixed-medium curriculum. However, further progression to a harder difficulty band reduces accuracy to 0.6405 ± 0.0053. These results show that staged difficulty progression can outperform persistent medium-difficulty selection, but its benefit is non-monotonic. Effective GRPO curricula can therefore benefit from progressing beyond medium difficulty, while avoiding excessive progression toward overly difficult samples.}


\maketitle

\section{Introduction}\label{sec:introduction}

Reinforcement learning with verifiable rewards (RLVR) provides a direct way to
improve reasoning models when outputs can be checked by explicit correctness rules.
Group Relative Policy Optimization (GRPO) constructs policy updates from multiple
responses to the same problem and does not require a separately trained value model
\cite{shao2024deepseekmath}. Under binary correctness rewards, problems whose
responses are nearly all correct or nearly all incorrect tend to produce low within-group
reward variance and thus weak group-relative learning signals
\cite{xu-etal-2026-prune}. In contrast, prompts whose sampled responses differ in
correctness provide nonzero reward variance for group-relative advantage estimation
and can therefore yield more informative learning signals \cite{le2025noprompt}.

This property raises important questions for curriculum learning in GRPO. Classical
curriculum learning typically organizes training samples from easy to hard according
to estimated difficulty \cite{bengio2009curriculum}. In GRPO, however, two practical
questions arise: whether difficulty estimates computed before training remain aligned
with the evolving policy, and whether progressively increasing curriculum difficulty
continues to provide useful learning signals. Because difficulty is relative to the current
policy, the same problem may move between easy, medium, and hard difficulty bands as
training proceeds. Moreover, a lower success rate does not necessarily correspond to a
more informative optimization signal. Consequently, a curriculum constructed from initial
difficulty can become temporally misaligned with the current policy. This motivates
estimating sample difficulty online under the current policy, so that training decisions
reflect the model's current state rather than stale initial estimates. However, online
difficulty estimation alone does not determine which difficulty band should be used as
the policy evolves.

\begin{figure}[!ht]
\centering
\includegraphics[width=\linewidth]{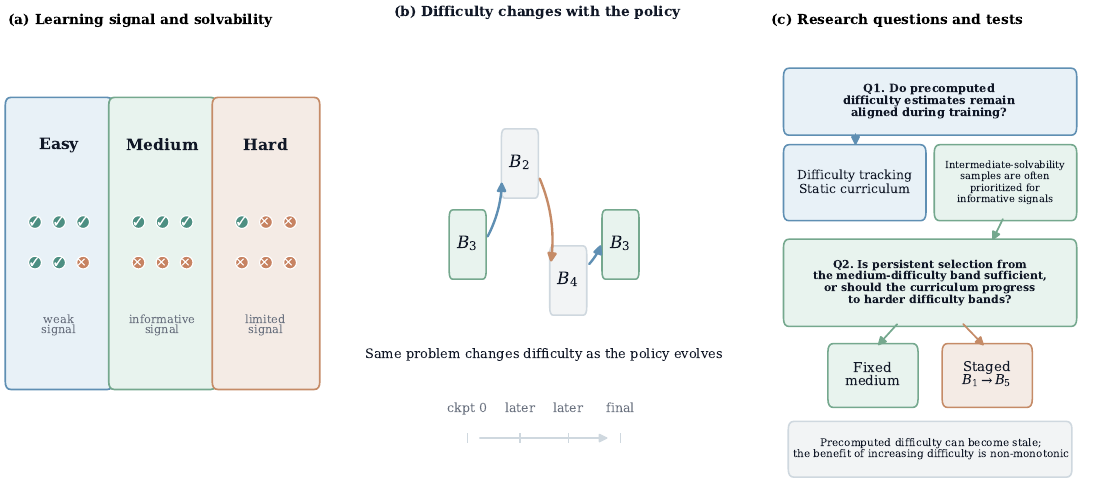}
\caption{Motivation and research questions for difficulty-aware training in
GRPO. (a) The usefulness of a training problem depends on its current
solvability: overly easy or hard problems tend to yield weak group-relative
learning signals, whereas intermediate-solvability problems are more likely to
produce informative comparisons among sampled responses.
(b) Sample difficulty depends on the current policy and can change across difficulty
bands as training proceeds, making precomputed difficulty estimates
potentially stale. (c) Intermediate-solvability samples are therefore natural
candidates for online selection and are emphasized by recent difficulty-aware
selection strategies. This motivates two questions examined in this study:
whether precomputed difficulty estimates remain aligned with the evolving
policy, and whether persistently selecting the medium-difficulty band is
sufficient or the curriculum should progress to harder difficulty bands during training.}
\label{fig:motivation}
\end{figure}

Figure~\ref{fig:motivation} summarizes the motivation of this work. In GRPO,
intermediate-solvability problems are attractive for online selection because
they are more likely to generate mixed correct and incorrect responses and
therefore informative group-relative signals. At the same time, difficulty may
change as the policy evolves, motivating our investigation of both stale
precomputed difficulty estimates and curriculum scheduling across difficulty bands.

Recent work has shown that adaptive curricula and online data selection can improve RLVR data efficiency by using reward statistics, estimated success rates, uncertainty, gradient information, or learnability signals \cite{adacurl,odf,dots,pcl,rftinf,insight,reschedule}. In GRPO, samples with intermediate success rates are particularly relevant because they are more likely to produce groups containing both correct and incorrect responses, thereby yielding nonzero group-relative learning signals \cite{xu-etal-2026-prune,le2025noprompt}. This raises a further question: is continually selecting such medium-difficulty samples sufficient to maximize final model performance, or can the model benefit from progressing toward harder samples as its policy evolves? If such progression is beneficial, it remains unclear whether the benefit persists as curriculum difficulty continues to increase.

We address these questions through a sequence of curriculum experiments. First, we track the same 1,495 GSM8K training samples throughout a continuous GRPO run to characterize how sample difficulty evolves with the policy. Second, we test whether a static curriculum constructed from initial difficulty becomes misaligned with the current policy. Third, we introduce a fixed-medium curriculum that keeps the active difficulty band fixed at $B_3$ while samples are reselected online according to the current policy. This setting tests whether continuously selecting samples from an informative medium-difficulty band is sufficient for maximizing final performance. Finally, we evaluate a staged curriculum in which the active difficulty band progresses from easier to harder bands according to the empirical success rate under the current policy. Across three random seeds, the staged curriculum achieves higher test performance at the moderately harder $B_4$ band than the fixed-medium curriculum. However, further progression into the harder $B_5$ band consistently reduces test performance. Together, these experiments suggest that medium-difficulty samples provide informative learning signals but are not sufficient by themselves to maximize final model capability. Progression to a moderately harder difficulty band can improve performance, whereas further increases in curriculum difficulty are not monotonically beneficial. Detailed metric definitions, additional training trajectories, and performance on the fixed diagnostic subsets used for checkpoint selection and stage transitions are provided in the appendices.

Figure~\ref{fig:framework} provides a visual comparison of the training
paradigms considered in this study, highlighting the differences between
static difficulty ordering, online sample selection, and
staged difficulty progression.

\begin{figure}[htbp]
\centering
\includegraphics[width=\linewidth]{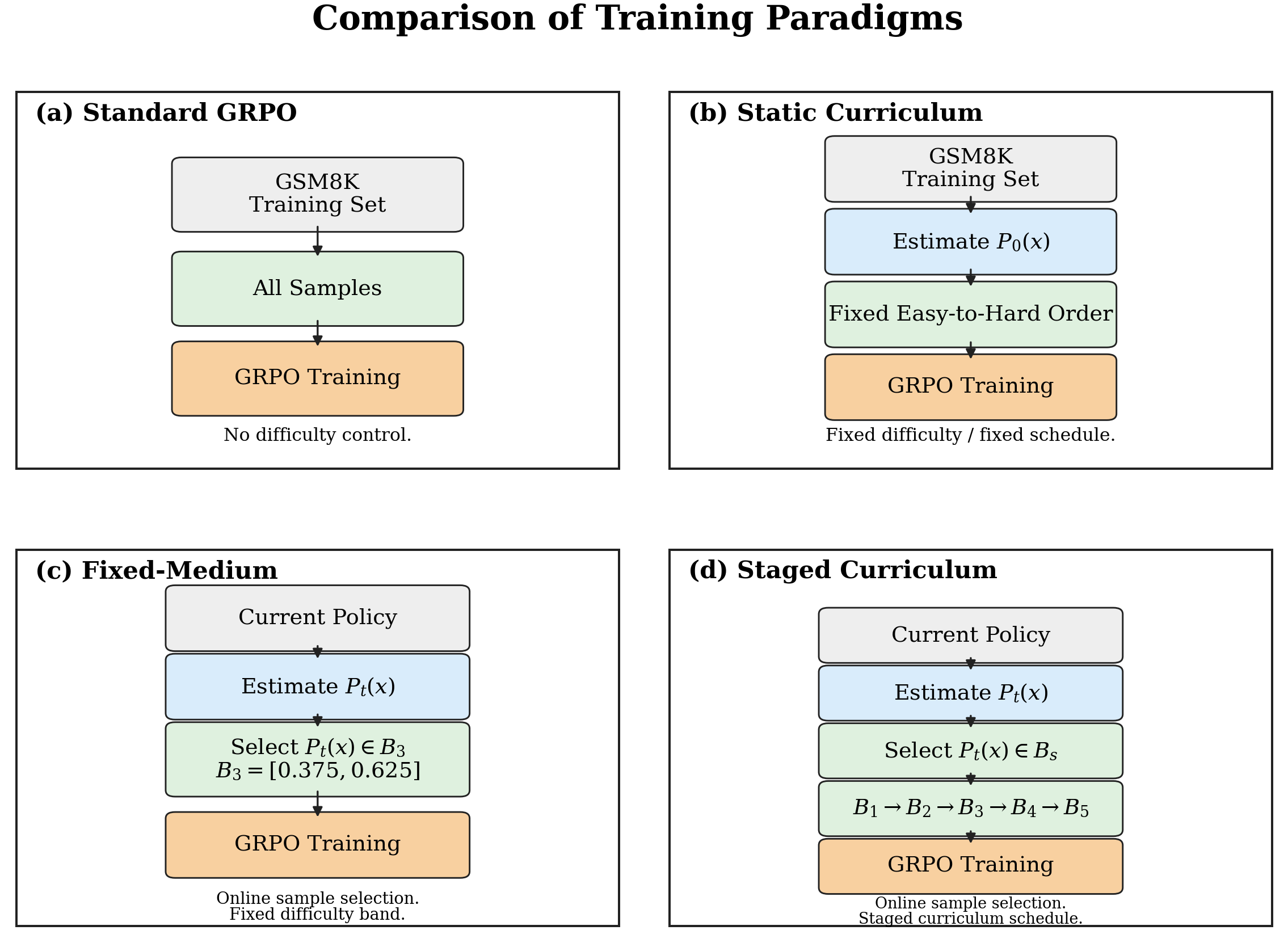}
\caption{Comparison of the training paradigms considered in this study.
Standard GRPO uses the full training set without difficulty-based selection.
The static curriculum uses a fixed easy-to-hard ordering determined by the
initial solvability estimate $P_0(x)$. The fixed-medium curriculum reselects samples
online according to the current-policy success rate while keeping difficulty band
$B_3$ fixed. The staged curriculum reselects samples online while moving the
active difficulty band through
$B_1 \rightarrow B_2 \rightarrow B_3 \rightarrow B_4 \rightarrow B_5$.}
\label{fig:framework}
\end{figure}

The main contributions of this work are summarized as follows:

\begin{itemize}
    \item \textbf{Policy-dependent difficulty dynamics.}
    We systematically characterize how sample difficulty changes during GRPO training and show that individual samples exhibit substantial and non-monotonic difficulty changes. We further observe that relatively stable global difficulty rankings can coexist with substantial turnover within specific difficulty bands, highlighting the limitations of relying on fixed difficulty estimates throughout training.

    \item \textbf{Staged difficulty progression.}
    We design and evaluate a staged curriculum that re-estimates sample difficulty under the current policy and progressively shifts the active difficulty band toward harder samples. Experiments across multiple random seeds show that persistent medium-difficulty selection does not achieve the strongest observed performance, while progressing to a moderately harder difficulty band provides additional gains. However, further progression to harder samples reduces performance, revealing a non-monotonic benefit of increasing curriculum difficulty.
\end{itemize}
The remainder of this paper is organized as follows.
Section~\ref{sec:related} reviews related work on GRPO and reinforcement learning with verifiable
rewards, dynamic data selection, and curriculum learning with model-dependent
difficulty.
Section~\ref{sec:framework} introduces the problem formulation and empirical framework, including
policy-dependent sample difficulty and the curriculum settings considered in this work.
Section~\ref{sec:setup} describes the experimental setup for difficulty tracking, static curriculum
training, the fixed-medium curriculum, and the staged curriculum.
Section~\ref{sec:results} presents the empirical results on difficulty dynamics, the misalignment of
precomputed difficulty, and performance under different curriculum conditions.
Section~\ref{sec:discussion} discusses the implications of the results, the relationship between online
sample selection and difficulty progression, as well as the limitations and future
directions of this study.
Finally, Section~\ref{sec:conclusion} concludes the paper by summarizing the main findings and their
implications for difficulty-aware GRPO training.
\section{Related Work}\label{sec:related}

\subsection{GRPO and RLVR}
RLVR has become a major approach for improving reasoning with rule-based or otherwise verifiable rewards. DeepSeekMath introduced GRPO for mathematical reasoning, and subsequent work extended verification-based reinforcement learning to broader reasoning tasks and formal theorem proving \cite{shao2024deepseekmath,guo2025deepseekr1,xin2025deepseekprover}. Compared with PPO-based approaches that rely on an additional value model, GRPO constructs relative advantages from multiple sampled responses within the same prompt, thereby reducing reliance on explicit value estimation. Recent studies have further investigated GRPO and related reinforcement learning frameworks from the perspectives of sampling strategy, objective design, and optimization stability \cite{dapo,drgrpo,gspo}.

In contrast to these algorithm-level improvements, our work focuses on the interaction between the evolving policy and training samples. Since GRPO derives learning signals from relative rewards within response groups, the usefulness of a problem is closely related to its current solvability under the current policy. However, existing GRPO studies mainly focus on optimization objectives and training stability, while the evolution of individual-sample difficulty with the policy has been less
systematically characterized.
\subsection{Dynamic Data Selection}
Dynamic data selection aims to improve training efficiency by identifying samples that are more valuable given the current model state. Offline selection methods estimate data value using reward trajectories, influence estimation, or multi-factor scoring \cite{limr,rftinf,depo}. In contrast, online selection methods adapt sample choices during optimization according to model-dependent signals, such as reward variance, predicted success probability, Bayesian uncertainty, transition dynamics, or gradient-related information \cite{odf,dots,mopps,dps,bots,uforl}. Other approaches, including PCL and CurES, further exploit intermediate success rates or gradient-efficient sampling strategies to improve training efficiency \cite{pcl,cures}. 

These studies motivate treating the utility of training samples as model-dependent rather than fixed. In particular, prioritizing samples with intermediate success rates provides a natural strategy for GRPO, since such samples are more likely to produce mixed correct and incorrect responses and therefore informative group-relative learning signals \cite{le2025noprompt,xu-etal-2026-prune}. Existing online selection methods already adapt sample choices according to the current model state. However, it remains unclear whether continuously selecting samples with intermediate solvability, even when sample difficulty is re-estimated online, is sufficient to maximize final model performance. Our work complements these studies by systematically characterizing how the difficulty
of the same samples evolves with the policy and by examining the sufficiency of persistent selection of
intermediate-solvability samples as the policy evolves.

\subsection{Curriculum Learning with Model-Dependent Difficulty}
Curriculum learning traditionally organizes training data according to estimated difficulty, following the principle of gradually increasing task complexity \cite{bengio2009curriculum}. Recent reasoning-oriented studies have incorporated curriculum strategies into reinforcement learning. E2H Reasoner explicitly organizes tasks in an easy-to-hard order \cite{e2hreasoner}, while Re-Schedule constructs curricula according to reasoning complexity rather than final-answer accuracy \cite{reschedule}. Other approaches, such as AdaRFT and AdaCuRL, further adjust training difficulty according to the evolving model state \cite{adarft,adacurl}.

These difficulty-aware curriculum approaches suggest that organizing training according to model-dependent difficulty can improve training efficiency by providing samples with appropriate learning challenges. However, traditional easy-to-hard curriculum learning is built on the intuition that models benefit from being exposed to increasingly complex examples as their capabilities improve \cite{bengio2009curriculum}. This assumption may not always hold in GRPO, where optimization depends on relative rewards among sampled responses. Extremely easy problems may provide limited learning signals due to low reward variance, while extremely difficult problems may also provide limited informative comparisons when most responses are incorrect \cite{xu-etal-2026-prune,le2025noprompt}.

Therefore, beyond estimating sample difficulty under the current policy, an important open question in GRPO curriculum learning is whether moving toward harder difficulty bands continues to provide benefits as the policy evolves. In particular, it remains unclear whether the gains from moving to harder difficulty bands persist as curriculum difficulty continues to increase. Our work investigates this issue by characterizing how difficulty evolves with the policy and evaluating a staged easy-to-hard curriculum across increasingly difficult bands. This allows us to examine whether progressively increasing curriculum difficulty remains beneficial, rather than assuming a monotonic benefit from moving to harder difficulty bands.
\FloatBarrier
\section{Problem Formulation and Empirical Framework}\label{sec:framework}

Tables~\ref{tab:notation_main} and~\ref{tab:notation_metrics}
summarize the main notation used throughout this paper.

\begin{table}[!t]
\centering
\caption{Main notation for policy-dependent difficulty and curriculum settings.}
\label{tab:notation_main}
\small
\begin{tabular}{p{0.23\linewidth} p{0.68\linewidth}}
\toprule
Symbol & Meaning \\
\midrule

$x$, $x_i$
& A training problem; $x_i$ denotes the $i$-th problem. \\

$y$
& A response sampled from the policy for problem $x$. \\

$t$, $t_k$
& Training state or checkpoint; $t_k$ denotes the $k$-th tracked checkpoint. \\

$\theta_t$, $\pi_{\theta_t}$
& Model parameters and the corresponding policy at training state $t$. \\

$V(x,y)$
& Binary correctness verifier for response $y$ to problem $x$. \\

$p_t(x)$
& True solvability of problem $x$ under the policy at training state $t$. \\

$d_t(x)$
& True policy-dependent difficulty, defined as $1-p_t(x)$. \\

$G$
& Number of sampled responses used to estimate problem solvability. \\

$r_{t,g}(x)$
& Binary correctness reward of the $g$-th sampled response for problem $x$
at training state $t$. \\

$P_t(x)$
& Empirical success rate estimated from $G$ responses sampled under the
policy at training state $t$. \\

$\widehat{d}_t(x)$
& Estimated policy-dependent difficulty, defined as $1-P_t(x)$. \\

$\mathcal{T}$
& Sequence of training checkpoints used to track difficulty changes. \\

$B_s=[l_s,u_s]$
& Difficulty band used at curriculum stage $s$, operationalized by the
success-rate interval $[l_s,u_s]$. \\

$B_1,\ldots,B_5$
& Predefined difficulty bands used in the staged curriculum. \\

$B_3$
& Medium-difficulty band kept fixed in the fixed-medium curriculum and used as
the third stage of the staged curriculum. \\

$a_t(x)$
& Binary gating variable indicating whether problem $x$ belongs to the active
difficulty band at training state $t$. \\

\bottomrule
\end{tabular}
\end{table}

\begin{table}[!t]
\centering
\caption{Notation for difficulty-dynamics and difficulty-band membership metrics.}
\label{tab:notation_metrics}
\small
\begin{tabular}{p{0.23\linewidth} p{0.68\linewidth}}
\toprule
Symbol & Meaning \\
\midrule

$R_i$
& Difference between the maximum and minimum empirical solvability of problem
$x_i$ over the tracked training trajectory. \\

$D_{i,\mathrm{init}}$
& Maximum absolute deviation of problem $x_i$ from its initial empirical
solvability over the tracked trajectory. \\

$\Delta_{i,k}$
& Difference in empirical solvability of problem $x_i$ between consecutive
checkpoints $t_{k-1}$ and $t_k$. \\

$\delta$
& Minimum-change threshold used to identify meaningful direction reversals. \\

$S_t(B)$
& Set of tracked problems whose empirical solvability lies in band $B$ at
training state $t$. \\

$N$
& Total number of problems in the fixed observation set. \\

$\mathrm{Retention}_t(B)$,
$\mathrm{Outflow}_t(B)$
& Fractions of initial band members that remain in or leave band $B$,
respectively. \\

$\mathrm{Inflow}_t(B)$
& Fraction of current members of band $B$ that were outside the band initially. \\

$\mathrm{Jaccard}_t(B)$
& Jaccard similarity between the initial and current membership of band $B$. \\

$\mathrm{Mismatch}_t(B)$
& Symmetric difference between initial and current band membership,
normalized by $N$. \\

$TV_i$
& Sum of the absolute empirical-solvability changes of problem $x_i$
across consecutive tracked checkpoints. \\

$E_i$
& Extra variation caused by back-and-forth changes in the empirical solvability
of problem $x_i$, defined as $TV_i-R_i$. \\

$\rho$, $\tau_b$
& Spearman and Kendall rank-correlation coefficients, respectively. \\

\bottomrule
\end{tabular}
\end{table}

\subsection{Difficulty Under the Current Policy}
Let $\pi_{\theta_t}$ denote the policy at training state $t$. For a problem $x$, we define its solvability under the current policy as
\begin{equation}
p_t(x)=\Pr_{y\sim\pi_{\theta_t}(\cdot|x)}[V(x,y)=1],
\end{equation}
where $V(x,y)\in\{0,1\}$ is the correctness verifier. The corresponding true difficulty is defined as $d_t(x)=1-p_t(x)$. With $G$ sampled responses, we estimate solvability as
\begin{equation}
P_t(x)=\frac{1}{G}\sum_{g=1}^{G} r_{t,g}(x),\qquad \widehat d_t(x)=1-P_t(x).
\end{equation}
Since the true solvability $p_t(x)$ cannot be computed exactly in practice, we use the empirical success rate $P_t(x)$ and the corresponding estimated difficulty $\hat d_t(x)$ for all analyses.

\subsection{Difficulty Dynamics}
Let $\mathcal{T}=\{t_0,t_1,\ldots,t_K\}$ denote the ordered sequence of checkpoints used for difficulty tracking.

We quantify the variation in empirical success rate for each problem over training using its training-period span
\begin{equation}
R_i=\max_{t\in\mathcal{T}}P_t(x_i)-\min_{t\in\mathcal{T}}P_t(x_i).
\end{equation}

We also measure the maximum absolute deviation from the initial empirical success rate as
\begin{equation}
D_{i,\mathrm{init}}
=
\max_{t\in\mathcal{T}}
\left|P_t(x_i)-P_0(x_i)\right|.
\end{equation}

To identify non-monotonic changes, we examine checkpoint-to-checkpoint differences
\begin{equation}
\Delta_{i,k}
=
P_{t_k}(x_i)-P_{t_{k-1}}(x_i).
\end{equation}

After discarding changes with $|\Delta_{i,k}|<\delta$, we count a direction reversal whenever two consecutive retained changes have opposite signs. Here, $\delta$ denotes the minimum-change threshold used to suppress small fluctuations caused by finite sampling. Since each empirical success-rate estimate is based on $G=32$ sampled responses, we set $\delta=0.125$ as the primary threshold, corresponding to four increments
of $1/32$ in the empirical success rate. Sensitivity analyses with alternative thresholds ($\delta=0.0625$ and $\delta=0.25$) are reported in Appendix~\ref{app:fullstats}.
We additionally examine global rank correlation and local membership changes in predefined success-rate intervals. Full definitions of retention, inflow, outflow, Jaccard similarity, mismatch, total variation, extra round-trip variation, and re-entry are given in Appendix~\ref{app:metrics}.

\subsection{Evaluation of Static Difficulty and Curriculum Scheduling}

We examine two related questions that arise because difficulty depends on the current
policy. First, we test whether curricula constructed from initial difficulty estimates
become misaligned with the current policy as training proceeds. Specifically, we
construct a static curriculum by estimating sample difficulty under the initial policy
and fixing the resulting difficulty ordering throughout training. This setting allows
us to examine whether the initial difficulty estimates remain aligned with difficulty
under the current policy.

Second, we study how curriculum difficulty should be scheduled once sample difficulty
is re-estimated online under the current policy. In particular, we ask whether
persistently using a medium-difficulty band is sufficient to maximize final performance
and whether progressing to harder difficulty bands continues to provide benefits as
curriculum difficulty increases. We use the term \emph{difficulty band} to denote a
predefined range of policy-dependent sample difficulty used for curriculum selection.
Each difficulty band $B_s$ is operationalized by a corresponding success-rate interval
$[l_s,u_s]$. At curriculum stage $s$, $B_s$ is referred to as the \emph{active difficulty
band}. A sample contributes to the policy loss only when its estimated solvability falls
within the success-rate interval associated with the active difficulty band:
\begin{equation}
a_t(x)=\mathbb{I}[P_t(x)\in B_s].
\end{equation}
Under the fixed-medium curriculum, the active difficulty band remains fixed at $B_3$
while eligible samples are reselected online as the policy evolves. Under the staged
curriculum, the active difficulty band progresses toward bands associated with lower
success rates and therefore greater difficulty. Comparing these curricula allows us to
examine whether online sample reselection is sufficient when the difficulty band remains
fixed and how performance changes as the curriculum progresses to harder difficulty bands.

This gating mechanism is used only to define the experimental conditions. An effective update is an optimizer step in which at least one sample satisfies the gate and a parameter update actually occurs.
\section{Experimental Setup}\label{sec:setup}

\subsection{Model and Training Configuration}

We use GSM8K \cite{cobbe2021gsm8k} with Qwen2.5-1.5B-Instruct \cite{qwen25technical} and LoRA fine-tuning \cite{hu2021lora}. Due to different computational requirements, we use different numbers of sampled responses for difficulty estimation and online optimization. Specifically, we use $G=32$ responses per problem for offline difficulty analysis and $G=8$ responses per problem during curriculum experiments. The main training runs use a batch size of 8, a gradient accumulation factor of 1,
a learning rate of $1\times10^{-5}$, a temperature of 0.7, top-$p$ of 0.85,
and a maximum generation length of 1,024 tokens. Answer correctness is used as the primary verifiable reward. An auxiliary length
reward with a fixed weight of 0.2 and a preferred response-length range of
60--650 tokens is also used, but it is not used to compute $P_t(x)$ or make gating
decisions. The official GSM8K test set is used only for generalization evaluation.

\begin{table}[t]
\centering
\caption{Difficulty bands and corresponding success-rate intervals used in the online curriculum conditions.}
\label{tab:bands}
\begin{tabular}{cccc}
\toprule
Difficulty band & Success-rate interval & Correct out of 8 & Relative difficulty \\
\midrule
$B_1$ & $[0.625,0.875]$ & 5--7 & Easier \\
$B_2$ & $[0.500,0.750]$ & 4--6 & Moderately easy \\
$B_3$ & $[0.375,0.625]$ & 3--5 & Medium \\
$B_4$ & $[0.250,0.500]$ & 2--4 & Moderately hard \\
$B_5$ & $[0.125,0.375]$ & 1--3 & Harder \\
\bottomrule
\end{tabular}
\end{table}

\subsection{Difficulty Tracking and Static Curriculum}

Before training, all 7,473 GSM8K training problems are evaluated with 32 generations each to obtain their initial solvability estimates. We first perform a reference GRPO run using the standard GRPO objective without difficulty-based filtering or curriculum scheduling. During this training process, checkpoints are saved at $\{0,10,20,30,40,50,100,250,500,1000,1500,2000,4000,7473\}$.

For performance comparison, we independently run the standard GRPO baseline with the same three random seeds used in the curriculum experiments. Each run is trained for one complete pass over the 7,473 GSM8K training problems. We retrospectively report the best observed test accuracy along each completed training trajectory; these test values are used only for post-hoc performance comparison and do not affect training, early stopping, or parameter updates.

A fixed stratified observation set of 1,495 problems (20\% of the training set) is constructed by proportional sampling based on the initial solvability distribution $P_0(x)$, preserving the original difficulty distribution of the full training set. At each checkpoint, the observation set is reevaluated with 32 generations under the corresponding policy to obtain updated solvability estimates $P_t(x)$ and characterize difficulty changes over training.

The static easy-to-hard curriculum first estimates the difficulty of training samples
under the initial policy and ranks all training problems in descending order based on
their initial solvability $P_0(x)$, where a higher solvability score indicates an easier
sample. This ordering is fixed throughout training without further difficulty
re-estimation. To assess how the fixed difficulty ordering becomes misaligned with the
evolving policy, the next 500 samples in the training order are reevaluated at steps
1000, 2500, 3500, 4500, 5500, and 6500.

The static curriculum is independently repeated with three random seeds. For the
multi-seed static curriculum comparison, we retrospectively report the best observed
test accuracy along each completed training trajectory; these test values are used only
for post-hoc performance comparison and do not affect training, early stopping, or
parameter updates. Its best observed test performance is reported in Sec.~\ref{sec:static-curriculum-misalignment}, while
the complete difficulty-misalignment statistics for the static curriculum and multi-seed
results are provided in Appendix~\ref{app:additionaltraj}.

\subsection{Fixed-Medium and Staged Curricula}

The staged curriculum progressively moves the active difficulty band through
$B_1 \rightarrow B_2 \rightarrow B_3 \rightarrow B_4 \rightarrow B_5$.
Based on the initial solvability estimate $P_0(x)$, five fixed diagnostic
subsets are constructed sequentially for $B_1$ through $B_5$, with 128
problems assigned to each subset. To ensure that the diagnostic subsets are
mutually exclusive despite the overlapping success-rate intervals, any problem
already assigned to an earlier diagnostic subset is excluded from subsequent
subset construction. These diagnostic subsets are used only to monitor the
staged curriculum and are excluded from training, leaving 6,833 training
candidates.

Every 50 effective updates, the diagnostic subset corresponding to the current
curriculum stage is evaluated, with eight responses generated for each problem.
Stage transitions follow a fixed patience-based rule applied to the diagnostic
trajectory. A later checkpoint is treated as a new best only when its diagnostic
accuracy exceeds the current best by more than the empirical noise tolerance of
0.0107. Once a new best checkpoint is established, training continues for two
additional diagnostic evaluations after 50 and 100 additional effective updates,
respectively. If neither subsequent evaluation improves upon the current best by
more than 0.0107, the stage is considered to have reached a plateau. Neither fluctuations within the noise tolerance nor performance decreases are treated as meaningful improvements.

If a new best is observed during the patience window, the 100-update patience
criterion is restarted from that checkpoint. The best checkpoint retained for
the current stage is then used to initialize training in the next difficulty
band. The GSM8K test set is used only for generalization evaluation and is
never involved in stage-transition or checkpoint-selection decisions. The full
diagnostic trajectories, noise analysis, and selected checkpoints are reported
in Appendix~\ref{app:e4diagnostic}.

Motivated by online data-selection strategies that prioritize samples with intermediate
success rates, we introduce a fixed-medium curriculum for complementary comparison.
Such samples are attractive in GRPO because they are more likely to produce mixed
correct and incorrect responses and therefore informative group-relative learning signals.
However, informative learning signals do not necessarily translate into the strongest
final model performance. We therefore ask whether repeatedly selecting the medium-difficulty band is sufficient to maximize final model performance.

In the fixed-medium curriculum, the active difficulty band remains fixed at
$B_3=[0.375,0.625]$ throughout optimization, while samples are reselected online
according to the evolving policy. This curriculum preserves online difficulty estimation
and online sample selection while keeping the active difficulty band fixed.
Comparing fixed-medium curriculum with the staged curriculum allows us to examine
whether progressing to harder difficulty bands can provide additional benefits beyond
repeatedly selecting samples currently in the medium-difficulty band.

Because the fixed-medium curriculum has no stage-transition criterion, we report the
best observed test accuracy along each trajectory for comparison; these test values
are not used to guide training or parameter updates. This retrospective best-observed
value is used only for post-hoc comparison with the staged curriculum and does not
affect training or staged checkpoint selection.

All multi-seed experiments use the same three random seeds. For reporting convenience,
we refer to them as Seed 1, Seed 2, and Seed 3, corresponding to random seeds
42, 2026, and 100, respectively. For each staged run, stage transitions and checkpoint
selection follow the same diagnostic-based criterion described above. We compare
performance at the selected $B_4$ checkpoint with performance after 100 additional
effective updates in $B_5$. Multi-seed results are reported as mean $\pm$ standard
deviation.
\FloatBarrier
\section{Results}\label{sec:results}

\subsection{State Dependence and Non-Monotonicity of Difficulty}
The 1,495 tracked problems become easier on average, with the mean empirical success
rate increasing from 0.6554 to 0.7975. At the same time, individual trajectories vary
substantially. The mean training-period span is 0.2770, and 54.72\% of samples have
a training-period span of at least 0.25. At the final checkpoint, the mean absolute change from the initial empirical success
rate $P_0$ is 0.1583, with 50.50\% of samples changing by at least 0.125 relative
to their initial values.

Non-monotonic changes are common. With $\delta=0.125$, 53.38\% of samples exhibit
at least one direction reversal, and 36.79\% exhibit at least two. The prevalence of
direction reversals is threshold-dependent, rising to 87.02\% for $\delta=0.0625$ and
falling to 4.95\% for $\delta=0.25$. Across the five difficulty bands, roughly 81--84\%
of samples that leave their initial band later re-enter it. Detailed sensitivity and
re-entry statistics are reported in Appendix~\ref{app:fullstats}.

Figure~\ref{fig:difficulty_drift} summarizes both the population-level magnitude and non-monotonicity of
changes in empirical solvability, while also providing representative problem-level
trajectories.

\begin{figure}[!htbp]
\centering
\includegraphics[width=0.95\linewidth]{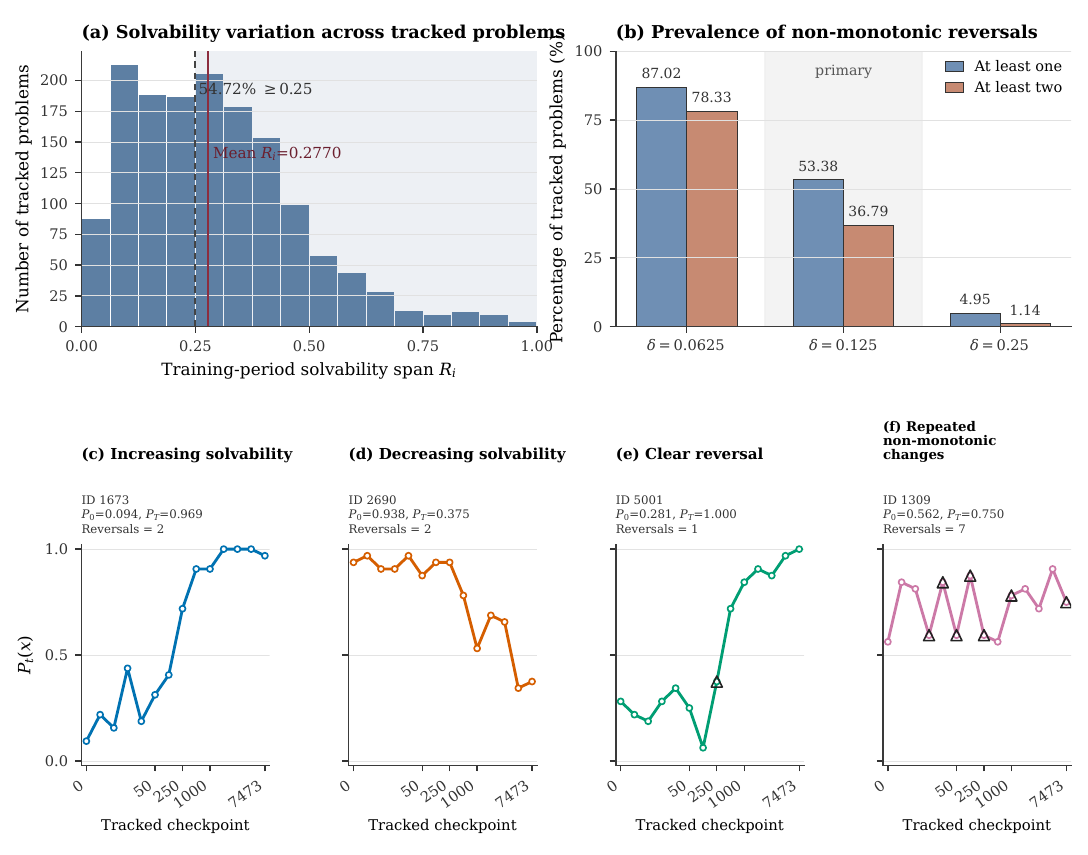}
\caption{
Magnitude and non-monotonicity of changes in empirical solvability during GRPO training.
(a) Distribution of the training-period solvability span $R_i$ across the 1,495 tracked GSM8K problems.
The dashed line marks $R_i=0.25$; 54.72\% of samples have a span of at least 0.25, and the mean span is 0.2770.
(b) Fractions of tracked problems exhibiting at least one and at least two direction reversals under different minimum-change thresholds $\delta$.
At the primary threshold $\delta=0.125$, 53.38\% of samples exhibit at least one reversal, and 36.79\% exhibit at least two.
(c)--(f) Representative trajectories illustrating increasing solvability, decreasing solvability, a clear direction reversal, and repeated non-monotonic changes, respectively.
The label above each panel reports the sample ID, initial solvability $P_0$, final solvability $P_T$, and the number of direction reversals.
For the representative trajectories, tracked checkpoints are shown at equal horizontal spacing for readability; tick labels indicate the corresponding training steps.
}
\label{fig:difficulty_drift}
\end{figure}

\begin{table}[htbp]
\centering
\caption{Core evidence for state-dependent difficulty dynamics.}
\label{tab:core-drift}
\begin{tabular}{lc}
\toprule
Statistic & Result \\
\midrule
Mean training-period span $R_i$ & 0.2770 \\
Samples with $R_i\geq0.25$ & 54.72\% \\
At least one direction reversal, $\delta=0.125$ & 53.38\% \\
Final Spearman correlation with $P_0$ & 0.8092 \\
Final $B_3$ retention & 20.74\% \\
Final $B_3$ Jaccard similarity & 0.1497 \\
\bottomrule
\end{tabular}
\end{table}

\subsection{Global Ordering Persistence and Local Band-Membership Change}
Global rank ordering remains largely preserved even when local band membership changes substantially. The Spearman rank correlation between $P_0$ and the final
$P_T$ remains 0.8092. In contrast, only 20.74\% of samples initially in the medium-difficulty
band $B_3=[0.375,0.625]$ remain there at the final checkpoint, while 65.00\% of
the samples in $B_3$ at the final checkpoint entered from outside the initial band.
These results show that a relatively stable global difficulty ranking does not imply
stable membership within a specific difficulty band. In other words, rank correlation
can remain high even when the set of samples occupying a given difficulty band changes
substantially. Full rank and membership tables are given in Appendix~\ref{app:fullstats}. An additional
visualization of the relationship between the initial and final rank percentiles is provided
in Appendix~\ref{app:rank-consistency-visualization}, where the concentration near the diagonal offers a more direct view
of the preserved global ordering.

\begin{figure}[htbp]
\centering
\includegraphics[width=\linewidth]{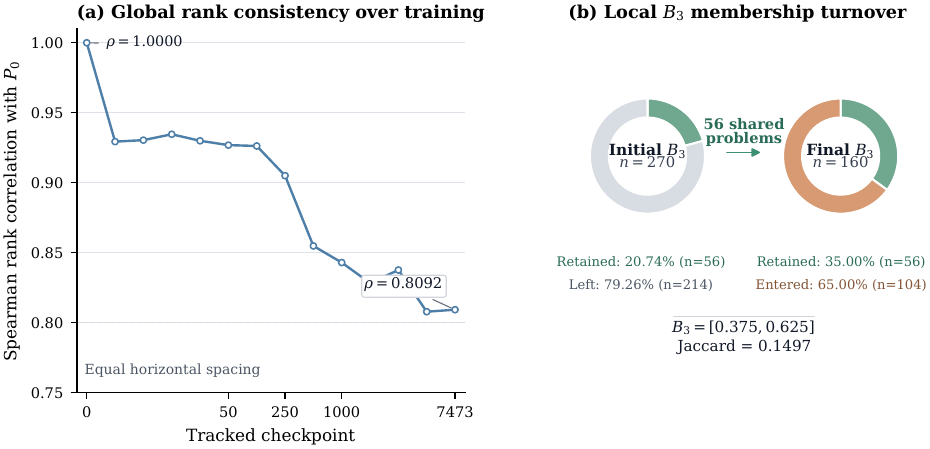}
\caption{
Global rank consistency and membership turnover in the medium-difficulty band.
(a) Spearman rank correlation between the initial solvability ranking and the ranking
at each tracked checkpoint for the 1,495 tracked problems. Although rank consistency
gradually decreases during training, it remains relatively high at the final checkpoint
($\rho=0.8092$). Tracked checkpoints are shown at equal horizontal spacing for
readability; tick labels indicate the corresponding training steps.
(b) Membership turnover in the medium-difficulty band
$B_3=[0.375,0.625]$. Of the 270 problems initially in $B_3$, only 56 remain in the
final $B_3$, corresponding to 20.74\% retention. The final $B_3$ contains 160
problems, of which 104 entered from outside the initial band, corresponding to
65\% inflow. The Jaccard similarity between the initial and final $B_3$ sets is
0.1497.
}
\label{fig:global-local-drift}
\end{figure}

Figure~\ref{fig:global-local-drift} visualizes this contrast directly: global rank consistency remains relatively
high throughout training, whereas membership in the medium-difficulty band
undergoes substantial turnover.

\subsection{Precomputed Difficulty Misalignment in Static Curriculum Training}\label{sec:static-curriculum-misalignment}

The static curriculum provides evidence of difficulty misalignment over training.
At step 5500, the next 500 scheduled problems have a planned mean success rate
of 0.3986 under the initial policy but a current mean success rate of 0.5573,
with a mean absolute deviation of 0.1863. Similar mismatches appear at earlier
checkpoints, although their magnitudes are not monotonically related to test accuracy.
We therefore interpret this analysis as evidence that precomputed difficulty estimates
can become temporally misaligned with the evolving policy, rather than as evidence
that such misalignment alone causes performance degradation. The complete table is
reported in Appendix~\ref{app:additionaltraj}.

Despite this misalignment, static curriculum training remains a competitive baseline.
Across Seeds 1, 2, and 3, the best observed test accuracies are 0.6535, 0.6497,
and 0.6535, respectively, corresponding to $0.6522 \pm 0.0022$. Therefore, the
observed difficulty mismatch should not be interpreted as implying that a precomputed
easy-to-hard curriculum becomes ineffective. Rather, these results indicate that the difficulty estimates underlying the fixed
schedule can become outdated even when the resulting curriculum still achieves
strong performance.
\FloatBarrier
\subsection{Performance Across Difficulty Bands}\label{sec:curriculum-difficulty-levels}

The two online curriculum conditions highlight two complementary effects. First, in
one detailed experimental run, persistently selecting the medium-difficulty band
does not yield the strongest observed performance. The fixed-medium curriculum reaches a
best observed test accuracy of 0.6459 while keeping the difficulty band at
$B_3=[0.375,0.625]$, whereas the staged curriculum reaches 0.6558 at the selected
$B_4$ checkpoint. This comparison suggests that repeatedly selecting samples
currently in the medium-difficulty band, despite their informative group-relative
learning signals, is not sufficient to achieve the strongest observed performance;
moving to a moderately harder difficulty band can provide additional gains.

Second, the benefit of progression does not persist as curriculum difficulty continues to
increase. After the staged run transitions from $B_4$ to the harder $B_5$ band,
test accuracy decreases from 0.6558 to 0.6391 after 50 additional effective updates
and further to 0.6353 after 100 additional effective updates. The corresponding $B_5$ diagnostic accuracy
also decreases from 0.5215 at entry to 0.4971 and 0.4512. The concurrent decline in
both diagnostic and test performance suggests that this deterioration is not merely
a transient fluctuation in test accuracy.

Taken together, this detailed run indicates that the performance benefit of moving to
harder difficulty bands is non-monotonic: remaining in the medium-difficulty band
can leave additional gains unrealized, whereas further progression into the harder
difficulty band can offset the gains obtained from moderate progression.

Although the staged and fixed-medium curricula follow different training trajectories,
the observed performance gap suggests that persistently selecting the medium-difficulty
band is insufficient to achieve the strongest observed test performance.
Sec.~\ref{sec:robustness-seeds} examines the consistency of this pattern across random seeds.

Figure~\ref{fig:performance-comparison}(a) visualizes the Seed 1 trajectory,
including the improvement observed during $B_4$ and the subsequent decline
after progression into $B_5$.

\FloatBarrier
\begin{table}[htbp]
\centering
\caption{Single-run outcomes for the main curriculum conditions.}
\label{tab:single_run_online}
\begin{tabular}{lll}
\hline
Setting & Evaluation point & Test accuracy \\
\hline
Fixed medium & Best observed checkpoint & 0.6459 \\
Staged curriculum & Selected $B_4$ checkpoint & 0.6558 \\
Staged curriculum & $B_5$ + 50 updates & 0.6391 \\
Staged curriculum & $B_5$ + 100 updates & 0.6353 \\
\hline
\end{tabular}
\end{table}

\FloatBarrier
\subsection{Robustness Across Random Seeds}\label{sec:robustness-seeds}

The qualitative pattern described in Sec.~\ref{sec:curriculum-difficulty-levels} is observed across all three random
seeds. As shown in Table~\ref{tab:multiseed_online}, at the selected $B_4$ checkpoints, the staged curriculum
achieves test accuracies of 0.6558, 0.6535, and 0.6679 for Seeds 1, 2, and 3,
respectively, corresponding to $0.6591 \pm 0.0077$. In comparison, the best observed
accuracies along the fixed-medium trajectories are 0.6459, 0.6422, and 0.6467,
yielding $0.6449 \pm 0.0024$. Performance at the selected $B_4$ checkpoints exceeds
the best observed fixed-medium performance in all three runs by 0.99, 1.13, and
2.12 percentage points, respectively.

For the standard GRPO baseline, the best observed test accuracies are 0.6459, 0.6414,
and 0.6459 for Seeds 1, 2, and 3, respectively, corresponding to
$0.6444 \pm 0.0026$. Performance at the selected $B_4$ checkpoints of the staged
curriculum is higher than the best observed standard GRPO performance in all three
runs, by 0.99, 1.21, and 2.20 percentage points, respectively.

\begin{table}[htbp]
\centering
\caption{Multi-seed performance of the curriculum conditions and baselines.}
\label{tab:multiseed_online}
\begin{tabular}{lcccc}
\hline
Setting & Seed 1 & Seed 2 & Seed 3 & Mean $\pm$ Std. \\
\hline
Standard GRPO: best observed
& 0.6459 & 0.6414 & 0.6459 & $0.6444 \pm 0.0026$ \\

Static curriculum: best observed
& 0.6535 & 0.6497 & 0.6535 & $0.6522 \pm 0.0022$ \\

Fixed medium: best observed
& 0.6459 & 0.6422 & 0.6467 & $0.6449 \pm 0.0024$ \\

Staged: selected $B_4$
& 0.6558 & 0.6535 & 0.6679 & $0.6591 \pm 0.0077$ \\

Staged: $B_5$ + 100
& 0.6353 & 0.6404 & 0.6458 & $0.6405 \pm 0.0053$ \\
\hline
\end{tabular}
\end{table}

\begin{figure}[t]
\centering
\includegraphics[width=\linewidth]{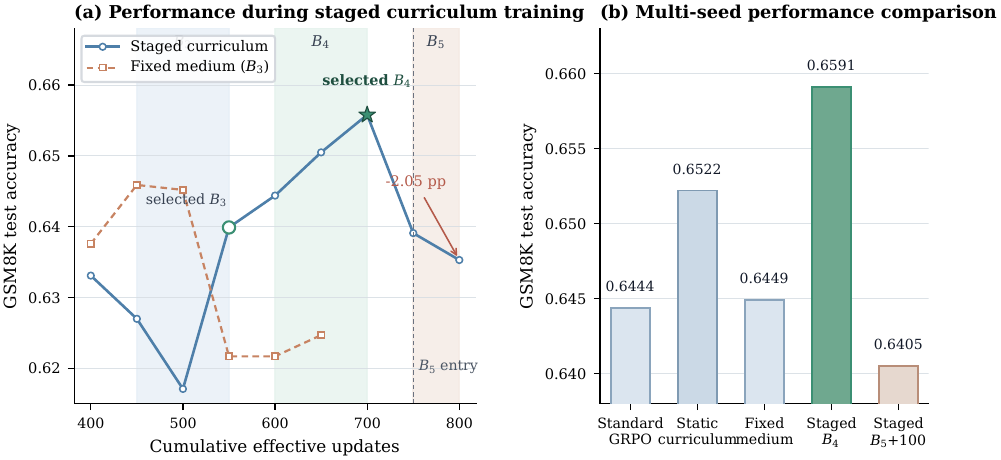}
\caption{Performance under the fixed-medium curriculum and the staged curriculum.
(a) Seed 1 test-accuracy trajectories for the fixed-medium curriculum and the staged
curriculum. The staged curriculum improves after progression into $B_4$,
reaching the selected $B_4$ checkpoint at 0.6558, whereas further progression
into $B_5$ reduces accuracy to 0.6353 after 100 additional effective updates.
Shaded areas indicate the active difficulty band and are shown only to
visualize stage progression.
(b) Multi-seed comparison of standard GRPO, static curriculum, fixed-medium
training, the selected staged $B_4$ checkpoint, and performance after 100
additional effective updates in $B_5$. The bar chart shows the mean GSM8K
test accuracy for each training setting to provide a direct visual comparison
of overall performance. Detailed per-seed values are reported in Table~\ref{tab:multiseed_online}.}
\label{fig:performance-comparison}
\end{figure}

Figure~\ref{fig:performance-comparison}(b) summarizes the corresponding multi-seed comparison across the main
training conditions. The static curriculum provides an additional strong baseline,
with best observed test accuracies of 0.6535, 0.6497, and 0.6535 across the three
runs, corresponding to $0.6522 \pm 0.0022$. Nevertheless, performance at the
selected $B_4$ checkpoints is higher than that of the static curriculum in all three
runs by 0.23, 0.38, and 1.44 percentage points, respectively.

Further progression into $B_5$ shows the opposite pattern across all three seeds.
After 100 additional effective updates in $B_5$, test accuracies decrease to 0.6353,
0.6404, and 0.6458 for Seeds 1, 2, and 3, respectively, corresponding to
$0.6405 \pm 0.0053$. Relative to the selected $B_4$ checkpoints, these results
represent decreases of 2.05, 1.31, and 2.21 percentage points, respectively.

Taken together, the multi-seed results support the same conclusion as the detailed
analysis in Sec.~\ref{sec:curriculum-difficulty-levels}: persistent selection of the medium-difficulty band is
insufficient to attain the strongest observed test performance, while test performance
at the selected $B_4$ checkpoints is consistently higher than the best observed
results from standard GRPO, the fixed-medium curriculum, and the static curriculum
across all three seeds. However, further progression into $B_5$ consistently reduces
performance. These results are consistent with a non-monotonic relationship between
scheduled difficulty and performance, rather than a monotonic benefit from continually
increasing difficulty.

\section{Discussion}
\label{sec:discussion}

Our results provide two main insights into difficulty-aware GRPO training. First,
sample difficulty is policy-dependent and can change substantially and
non-monotonically as training proceeds. Consequently, difficulty estimates obtained
under the initial policy may become misaligned with the current policy,
even when the global difficulty ranking remains relatively stable. The competitive
performance of the static curriculum, however, shows that such misalignment does
not by itself make a precomputed curriculum ineffective. Rather, these observations
motivate re-estimating sample difficulty under the current policy when sample
selection depends on current solvability.

More importantly, our experiments distinguish informative learning signals from
final model performance. Medium-difficulty samples are attractive for online
selection because they are more likely to produce mixed correct and incorrect
responses and therefore informative group-relative learning signals. However, the
fixed-medium curriculum shows that persistently selecting such samples does not
yield the strongest observed performance. The staged curriculum instead benefits
from progressing to a moderately harder difficulty band, whereas further progression
to a harder difficulty band consistently reduces performance across random seeds. This pattern
suggests a non-monotonic relationship between curriculum difficulty and model
performance: remaining at medium difficulty can leave additional gains unrealized,
while increasing difficulty too far can offset the gains obtained from moderate
progression.

These findings suggest that online sample selection and difficulty progression
should be treated as two related but distinct components of difficulty-aware GRPO
training. Online difficulty estimation determines which samples currently belong to
a given difficulty band, whereas difficulty progression determines which difficulty
band is used at different stages of training. Our results indicate that combining
online re-estimation with staged difficulty progression can improve performance
beyond persistent medium-difficulty selection, but progression toward harder samples
should not be continued indefinitely. Effective curriculum design therefore requires
not only identifying informative samples under the current policy, but also controlling
how far the curriculum progresses in difficulty.

Several limitations remain. Our experiments focus on a single model family and the
GSM8K benchmark, so the generality of the observed difficulty dynamics and
curriculum pattern should be evaluated across broader models and reasoning tasks.
Online solvability is estimated from only eight sampled responses, which introduces
noise into difficulty estimation and sample selection. In addition, the staged
curriculum relies on fixed diagnostic subsets and a predefined transition criterion,
and it uses a smaller training candidate pool than the other conditions. Comparisons
involving the staged curriculum should therefore be interpreted as
complete-procedure comparisons rather than strict single-factor evaluations. Future
work should examine these findings in broader settings, improve the accuracy and
efficiency of online difficulty estimation, and develop more general adaptive
mechanisms for determining when curriculum difficulty should increase and when
further progression should stop.

\section{Conclusion}\label{sec:conclusion}

We investigated how sample difficulty changes during GRPO training and whether
persistently selecting medium-difficulty samples is sufficient for achieving strong
final performance. The results show that sample difficulty changes with the evolving
policy, supporting the use of current-policy difficulty estimates for online sample
selection.

More importantly, the curriculum experiments reveal that persistent
medium-difficulty selection does not achieve the strongest observed performance.
Across three random seeds, progressing to a moderately harder difficulty band
provides additional gains, whereas further progression to a harder difficulty band consistently
reduces performance. These findings show that the benefit of increasing curriculum
difficulty is non-monotonic. Overall, effective GRPO curriculum design should combine
online sample selection with staged difficulty progression, moving beyond persistent
medium-difficulty selection while avoiding excessive progression toward overly
difficult samples.
\backmatter

\bmhead{Acknowledgements}

The authors would like to thank ITREC DMU for providing the computational server resources used in this study.

\section*{Declarations}

\subsection*{Funding}

This research was funded by the basic scientific research project of the Department of Education of Liaoning Province, the People’s Republic of China, grant number 984230020.

\subsection*{Competing Interests}

The authors declare no competing interests.

\subsection*{Data Availability}

The GSM8K dataset used in this study is publicly available.

\subsection*{Code Availability}

The code used in this study is available at
\url{https://github.com/NHLling/GRPO-B}.

\subsection*{Author Contribution}

Ronghan Li contributed to the conceptualization, methodology, implementation,
experiments, data analysis, and writing of the original manuscript.
Fei Chen contributed to research supervision, discussion and refinement of the
research ideas and methodology, and review and revision of the manuscript.
Mingyu Lu contributed to review and revision of the manuscript.
All authors reviewed and approved the final manuscript.

\begin{appendices}
\renewcommand{\theHsection}{appendix.\Alph{section}}
\renewcommand{\theHtable}{appendix.\Alph{section}.\arabic{table}}
\renewcommand{\thefigure}{\Alph{section}\arabic{figure}}
\renewcommand{\theHfigure}{appendix.\Alph{section}.\arabic{figure}}
\renewcommand{\theHequation}{appendix.\Alph{section}.\arabic{equation}}

\section{Detailed Difficulty-Dynamics Metrics}\label{app:metrics}

This appendix gives the detailed metrics used to characterize local difficulty-band changes and non-monotonic trajectories.

For a difficulty band $B$ with associated success-rate interval $[l,u]$, let
\begin{equation}
S_t(B)=\{x_i\mid P_t(x_i)\in [l,u]\}.
\end{equation}
We characterize changes in band membership using the following complementary metrics.
\begin{align}
\mathrm{Retention}_t(B)&=\frac{|S_0(B)\cap S_t(B)|}{|S_0(B)|},\\
\mathrm{Outflow}_t(B)&=1-\mathrm{Retention}_t(B),\\
\mathrm{Inflow}_t(B)&=\frac{|S_t(B)\setminus S_0(B)|}{|S_t(B)|},\\
\mathrm{Jaccard}_t(B)&=\frac{|S_0(B)\cap S_t(B)|}{|S_0(B)\cup S_t(B)|},\\
\mathrm{Mismatch}_t(B)&=\frac{|S_0(B)\triangle S_t(B)|}{N}.
\end{align}
Here $N$ is the number of samples in the fixed observation set and $\triangle$ denotes symmetric difference.

Retention measures the fraction of samples initially belonging to band $B$ that
remain in the same band at training state $t$, while Outflow measures the
complementary fraction that has left the band. Inflow measures the fraction of current band members that did not belong to $B$
initially, capturing replacement by newly entering samples. Jaccard similarity quantifies the overlap between the initial and current band memberships. Mismatch measures the size of their symmetric difference relative to the full observation set of N samples, reflecting the overall extent of membership disagreement. Together, these metrics characterize sample persistence, departure, entry, set overlap, and overall membership change.

To distinguish one-directional movement from repeated back-and-forth changes, we use
\begin{equation}
TV_i=\sum_{k=1}^{K}|P_{t_k}(x_i)-P_{t_{k-1}}(x_i)|,
\end{equation}
and define the extra round-trip variation as
\begin{equation}
E_i=TV_i-R_i.
\end{equation}
Here, $E_i=0$ for a monotonic trajectory, whereas $E_i>0$ indicates variation
beyond the observed success-rate span caused by back-and-forth changes.
To define re-entry, let
\begin{equation}
I_{i,k}(B)=\mathbf{1}\{P_{t_k}(x_i)\in B\}.
\label{eq:band-indicator}
\end{equation}
indicate whether sample $x_i$ belongs to band $B$ at checkpoint $t_k$.
A sample is counted as having ever left band $B$ if there exist checkpoints
$t_a<t_b$ such that $I_{i,a}(B)=1$ and $I_{i,b}(B)=0$. Among these samples,
a sample is counted as re-entering band $B$ if there exists a later checkpoint
$t_c>t_b$ such that $I_{i,c}(B)=1$. We define the re-entry rate as
\begin{equation}
\mathrm{ReEntry}(B)
=
\frac{
\left|\left\{x_i:\exists\,a<b<c,\;
I_{i,a}(B)=1,\;
I_{i,b}(B)=0,\;
I_{i,c}(B)=1
\right\}\right|
}{
\left|\left\{x_i:\exists\,a<b,\;
I_{i,a}(B)=1,\;
I_{i,b}(B)=0
\right\}\right|
}.
\end{equation}
Thus, the denominator consists only of samples that have been observed in band $B$ and
subsequently left it at least once, rather than all samples that ever belonged to
the band or the full observation set.

\section{Full Difficulty-Dynamics Statistics}\label{app:fullstats}

\begin{table}[htbp]
\centering
\caption{Magnitude of sample difficulty changes over the complete GRPO trajectory.}
\begin{minipage}{\linewidth}
\centering
\begin{tabular}{lccccc}
\toprule
Metric & Mean & Median & $\geq0.125$ & $\geq0.25$ & $\geq0.375$ \\
\midrule
Training-period span $R_i$ & 0.2770 & 0.2500 & 79.87\% & 54.72\% & 28.96\% \\
Maximum initial deviation $D_{i,\mathrm{init}}$ & 0.2207 & 0.1875 & 73.65\% & 37.19\% & 15.92\% \\
\bottomrule
\end{tabular}
\end{minipage}
\end{table}

\begin{table}[htbp]
\centering
\caption{Sensitivity of the direction-reversal rate to the minimum-change threshold.}
\begin{minipage}{\linewidth}
\centering
\begin{tabular}{ccc}
\toprule
Threshold $\delta$ & At least one reversal & At least two reversals \\
\midrule
0.0625 & 87.02\% & 78.33\% \\
0.125 & 53.38\% & 36.79\% \\
0.25 & 4.95\% & 1.14\% \\
\bottomrule
\end{tabular}
\end{minipage}
\end{table}

\begin{table}[htbp]
\centering
\caption{Rank consistency between the initial and current empirical success rates.}
\begin{minipage}{\linewidth}
\centering
\begin{tabular}{ccc}
\toprule
Training step & Spearman's $\rho$ & Kendall's $\tau_b$ \\
\midrule
100 & 0.9262 & 0.7988 \\
500 & 0.8548 & 0.7050 \\
1000 & 0.8429 & 0.6932 \\
7473 & 0.8092 & 0.6545 \\
\bottomrule
\end{tabular}
\end{minipage}
\end{table}

\begin{table}[htbp]
\centering
\caption{Membership changes in the medium-difficulty band $B_3=[0.375,0.625]$.}
\scriptsize
\setlength{\tabcolsep}{3pt}
\noindent\hfill
\begin{tabular}{ccccccc}
\toprule
Step & $|S_t|$ & Retention & Outflow & Inflow & Jaccard & Mismatch \\
\midrule
100 & 239 & 58.52\% & 41.48\% & 33.89\% & 0.4501 & 12.91\% \\
500 & 188 & 29.63\% & 70.37\% & 57.45\% & 0.2116 & 19.93\% \\
1000 & 187 & 27.04\% & 72.96\% & 60.96\% & 0.1901 & 20.80\% \\
7473 & 160 & 20.74\% & 79.26\% & 65.00\% & 0.1497 & 21.27\% \\
\bottomrule
\end{tabular}\hfill\mbox{}\par
\end{table}

\begin{table}[htbp]
\centering
\caption{Re-entry rates for the five difficulty bands.}
\noindent\hfill
\begin{tabular}{ccccc}
\toprule
Difficulty band & Ever in band & Ever left & Re-entered & Re-entry rate \\
\midrule
$B_1$ & 508 & 498 & 408 & 81.93\% \\
$B_2$ & 350 & 344 & 289 & 84.01\% \\
$B_3$ & 270 & 267 & 216 & 80.90\% \\
$B_4$ & 232 & 230 & 191 & 83.04\% \\
$B_5$ & 215 & 211 & 174 & 82.46\% \\
\bottomrule
\end{tabular}\hfill\mbox{}\par
\end{table}

\FloatBarrier
\clearpage
\subsection{Initial--Final Rank Consistency Visualization}
\label{app:rank-consistency-visualization}

\begin{figure}[htbp]
\centering
\includegraphics[width=0.68\linewidth]{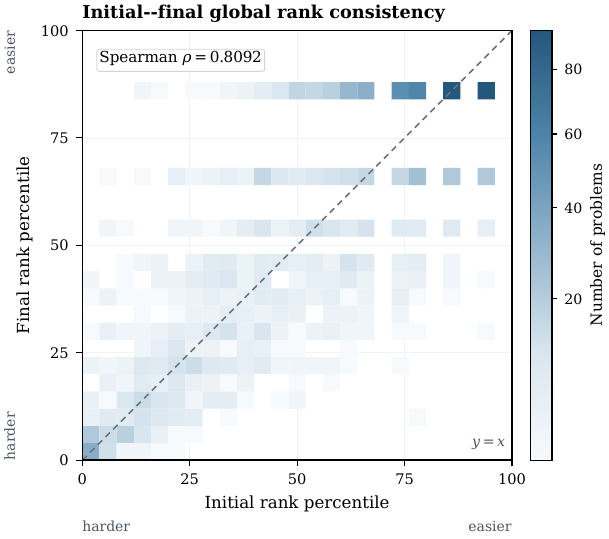}
\caption{
Initial--final global rank consistency for the 1,495 tracked GSM8K problems.
The axes show solvability-rank percentiles at the initial and final checkpoints,
with 0 indicating harder problems and 100 indicating easier problems; color denotes
the number of problems in each bin. The concentration near the dashed $y=x$ line
is consistent with the final Spearman rank correlation of $\rho=0.8092$.
}
\label{fig:rank_consistency_appendix}
\end{figure}

\FloatBarrier

\section{Diagnostic Trajectories and Stage Selection}
\label{app:e4diagnostic}

Five fixed diagnostic subsets of 128 problems are constructed for the staged
curriculum, one for each difficulty band $B_1$--$B_5$. The subsets are constructed
sequentially from $B_1$ through $B_5$, and any problem already assigned to an
earlier diagnostic subset is excluded from subsequent subset construction. These
subsets are used only to monitor training progress and are excluded from the training
pool, so they do not contribute to parameter updates. During each curriculum stage,
the diagnostic subset corresponding to the active difficulty band is evaluated every
50 effective updates, with eight responses generated for each problem.

To estimate the scale of diagnostic evaluation noise, the same model and diagnostic
subset are reevaluated ten times using different sampling seeds. The resulting
accuracies have a mean of 0.5226, a standard deviation of 0.0042, a minimum of
0.5156, and a maximum of 0.5283. The empirical 95th percentile of pairwise absolute accuracy differences is
approximately 0.0107. We use this value as an empirical noise tolerance
when determining whether a later diagnostic evaluation represents a meaningful
improvement over the current best checkpoint.

Stage transitions follow a fixed patience-based criterion. Within each stage, a later
checkpoint is treated as a new best checkpoint only when its diagnostic accuracy
exceeds the current best by more than the empirical noise tolerance of 0.0107.
Once a new best checkpoint is established, training continues for two additional
diagnostic evaluations after 50 and 100 additional effective updates, respectively.
If neither subsequent evaluation improves upon the current best by more than 0.0107,
the stage is considered to have reached a plateau. The retained best checkpoint is
then used to initialize training in the next difficulty band. If a new best checkpoint is
observed during this patience window, the 100-update patience criterion is restarted
from that checkpoint. When multiple checkpoints lie within the same high-performing
plateau and their diagnostic-accuracy differences remain within the empirical noise
tolerance, the earlier checkpoint is retained.

Table~\ref{tab:selected_stage_checkpoints} summarizes the checkpoints selected for the three random seeds.

\begin{table}[htbp]
\centering
\caption{Selected stage checkpoints for the three random seeds. Diagnostic accuracy
is measured on the fixed subset corresponding to the active difficulty band.}
\label{tab:selected_stage_checkpoints}
\small
\begin{tabular}{ccccc}
\toprule
Seed & Stage & Success-rate interval & Selected local update & Diagnostic accuracy \\
\midrule
1
& $B_1$ & $[0.625,0.875]$ & 150 & 0.8350 \\
1
& $B_2$ & $[0.500,0.750]$ & 250 & 0.8174 \\
1
& $B_3$ & $[0.375,0.625]$ & 150 & 0.7432 \\
1
& $B_4$ & $[0.250,0.500]$ & 150 & 0.6436 \\
\midrule
2
& $B_1$ & $[0.625,0.875]$ & 300 & 0.9111 \\
2
& $B_2$ & $[0.500,0.750]$ & 150 & 0.8232 \\
2
& $B_3$ & $[0.375,0.625]$ & 200 & 0.7715 \\
2
& $B_4$ & $[0.250,0.500]$ & 100 & 0.6172 \\
\midrule
3
& $B_1$ & $[0.625,0.875]$ & 250 & 0.9209 \\
3
& $B_2$ & $[0.500,0.750]$ & 300 & 0.8438 \\
3
& $B_3$ & $[0.375,0.625]$ & 50  & 0.7666 \\
3
& $B_4$ & $[0.250,0.500]$ & 100 & 0.6807 \\
\bottomrule
\end{tabular}
\end{table}

Figure~\ref{fig:diagnostic-trajectories} provides a visual summary of the diagnostic trajectories underlying the
stage-selection decisions reported in Tables~C6--C9.

\setcounter{figure}{0}
\begin{figure}[t]
\centering
\includegraphics[width=\linewidth]{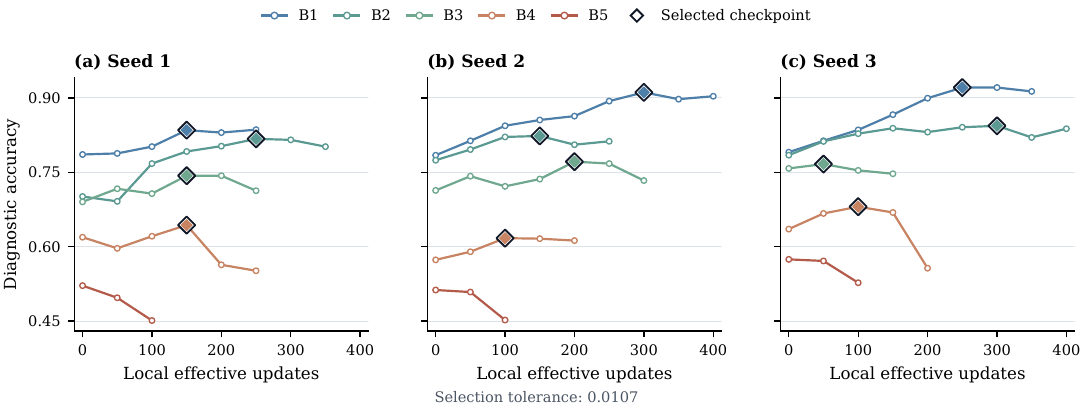}
\caption{
Diagnostic trajectories underlying stage selection across the three random seeds.
Each curve shows diagnostic accuracy as a function of local effective updates within
the corresponding difficulty band. Highlighted markers denote checkpoints retained
under the predefined patience-based criterion. A later checkpoint is considered a new
best only when its diagnostic accuracy exceeds the current best by more than the
empirical noise tolerance of 0.0107; after a new best is established, two additional
evaluations are performed after 50 and 100 additional effective updates, respectively.
No post-entry $B_5$ checkpoint is selected for any of the three seeds.
}
\label{fig:diagnostic-trajectories}
\end{figure}

The complete diagnostic
trajectories used for stage selection are reported below. Bold values indicate the
checkpoints retained for initialization of the subsequent stage.

\begin{table}[!ht]
\centering
\caption{Diagnostic trajectories for Seed 1. Columns denote local effective updates
within each difficulty band.}
\label{tab:diag_seed42}
\small
\setlength{\tabcolsep}{2pt}
\begin{tabular}{c@{\hspace{2pt}}ccccccccc}
\toprule
Stage & 0 & 50 & 100 & 150 & 200 & 250 & 300 & 350 & 400 \\
\midrule
$B_1$
& 0.7859 & 0.7881 & 0.8018 & \textbf{0.8350}
& 0.8301 & 0.8359 & -- & -- & -- \\

$B_2$
& 0.7012 & 0.6914 & 0.7676 & 0.7920
& 0.8027 & \textbf{0.8174} & 0.8154 & 0.8017 & -- \\

$B_3$
& 0.6904 & 0.7168 & 0.7070 & \textbf{0.7432}
& 0.7432 & 0.7129 & -- & -- & -- \\

$B_4$
& 0.6191 & 0.5967 & 0.6211 & \textbf{0.6436}
& 0.5635 & 0.5515 & -- & -- & -- \\

$B_5$
& 0.5215 & 0.4971 & 0.4512 & -- & -- & -- & -- & -- & -- \\
\bottomrule
\end{tabular}
\end{table}

\begin{table}[!ht]
\centering
\caption{Diagnostic trajectories for Seed 2. Columns denote local effective updates
within each difficulty band.}
\label{tab:diag_seed2026}
\small
\setlength{\tabcolsep}{2pt}
\begin{tabular}{c@{\hspace{2pt}}ccccccccc}
\toprule
Stage & 0 & 50 & 100 & 150 & 200 & 250 & 300 & 350 & 400 \\
\midrule
$B_1$
& 0.7844 & 0.8135 & 0.8438 & 0.8555
& 0.8633 & 0.8936 & \textbf{0.9111} & 0.8975 & 0.9033 \\

$B_2$
& 0.7743 & 0.7959 & 0.8213 & \textbf{0.8232}
& 0.8057 & 0.8125 & -- & -- & -- \\

$B_3$
& 0.7135 & 0.7422 & 0.7217 & 0.7363
& \textbf{0.7715} & 0.7676 & 0.7334 & -- & -- \\

$B_4$
& 0.5735 & 0.5898 & \textbf{0.6172} & 0.6162
& 0.6123 & -- & -- & -- & -- \\

$B_5$
& 0.5127 & 0.5086 & 0.4521 & -- & -- & -- & -- & -- & -- \\
\bottomrule
\end{tabular}
\end{table}

\begin{table}[!ht]
\centering
\caption{Diagnostic trajectories for Seed 3. Columns denote local effective updates
within each difficulty band.}
\label{tab:diag_seed100}
\small
\setlength{\tabcolsep}{2pt}
\begin{tabular}{c@{\hspace{2pt}}ccccccccc}
\toprule
Stage & 0 & 50 & 100 & 150 & 200 & 250 & 300 & 350 & 400 \\
\midrule
$B_1$
& 0.7906 & 0.8135 & 0.8354 & 0.8662
& 0.8994 & \textbf{0.9209} & 0.9209 & 0.9131 & -- \\

$B_2$
& 0.7845 & 0.8123 & 0.8279 & 0.8388
& 0.8311 & 0.8408 & \textbf{0.8438} & 0.8203 & 0.8379 \\

$B_3$
& 0.7578 & \textbf{0.7666} & 0.7539 & 0.7471
& -- & -- & -- & -- & -- \\

$B_4$
& 0.6355 & 0.6670 & \textbf{0.6807} & 0.6689
& 0.5566 & -- & -- & -- & -- \\

$B_5$
& 0.5745 & 0.5713 & 0.5273 & -- & -- & -- & -- & -- & -- \\
\bottomrule
\end{tabular}
\end{table}

No post-entry $B_5$ checkpoint is selected for any of the three seeds. For Seed~1,
the $B_5$ diagnostic accuracy decreases from 0.5215 at entry to 0.4971 after
50 effective updates and 0.4512 after 100 effective updates. For Seed~2, the
corresponding diagnostic accuracies are 0.5127, 0.5086, and 0.4521, while for
Seed~3 they are 0.5745, 0.5713, and 0.5273. Thus, no run shows a diagnostic
improvement after progression into $B_5$, and the selected $B_4$ checkpoint
remains the final selected checkpoint for each seed.

\section{Additional Training Trajectories}
\label{app:additionaltraj}

Table~\ref{tab:static_mismatch} reports the mismatch between the initial difficulty estimates used to
construct the static curriculum and the corresponding difficulty estimates under
the current policy at each evaluation point.

\begin{table}[htbp]
\centering
\caption{Difficulty mismatch between the static curriculum plan and the current
policy at each evaluation point.}
\label{tab:static_mismatch}
\begin{tabular}{cccc}
\toprule
Training step & Planned mean $P_0$ & Current mean $P_t$ & Mean $|P_t-P_0|$ \\
\midrule
1000 & 0.9531 & 0.9796 & 0.0396 \\
2500 & 0.8523 & 0.9422 & 0.1090 \\
3500 & 0.7528 & 0.8384 & 0.1562 \\
4500 & 0.6098 & 0.7275 & 0.1464 \\
5500 & 0.3986 & 0.5573 & 0.1863 \\
6500 & 0.1357 & 0.1544 & 0.0713 \\
\bottomrule
\end{tabular}
\end{table}

Figure~\ref{fig:static-mismatch} visualizes the difficulty misalignment summarized in Table~\ref{tab:static_mismatch}.

\setcounter{figure}{0}
\begin{figure}[t]
\centering
\includegraphics[width=\linewidth]{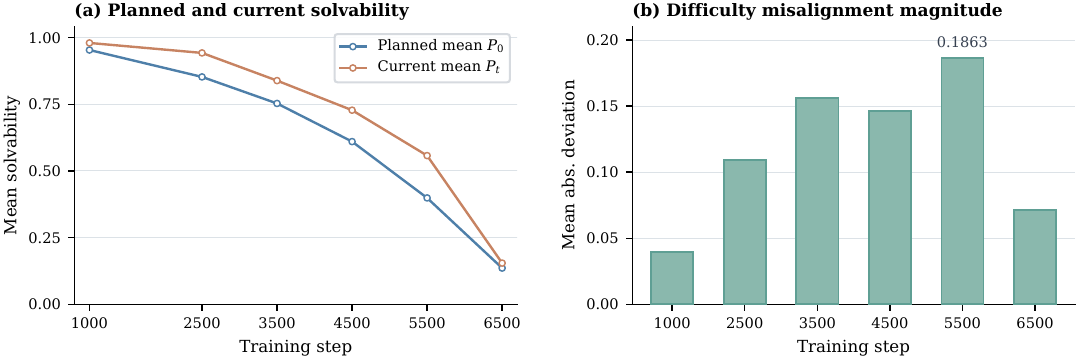}
\caption{
Difficulty misalignment in the static curriculum.
(a) Mean solvability specified by the precomputed curriculum schedule and mean solvability
under the current policy at each evaluation point.
(b) Mean absolute deviation between the planned and current mean solvability estimates
at the evaluated training steps.
}
\label{fig:static-mismatch}
\end{figure}

Despite the observed mismatch in precomputed difficulty estimates, the static
curriculum remains a competitive baseline across random seeds. 
Table~\ref{tab:static_multiseed} reports its best observed test performance across
random seeds.

\begin{table}[htbp]
\centering
\caption{Best observed test performance of the static curriculum across random seeds.}
\label{tab:static_multiseed}
\begin{tabular}{cccc}
\toprule
Seed 1 & Seed 2 & Seed 3 & Mean $\pm$ Std. \\
\midrule
0.6535 & 0.6497 & 0.6535 & $0.6522 \pm 0.0022$ \\
\bottomrule
\end{tabular}
\end{table}

Table~\ref{tab:staged_seed1_test} shows the test trajectory of the Seed~1 staged curriculum across
$B_3$, $B_4$, and $B_5$.

\begin{table}[htbp]
\centering
\caption{Seed 1 test trajectory for the staged curriculum.}
\label{tab:staged_seed1_test}
\begin{tabular}{ccl}
\toprule
Effective updates & Test accuracy & Curriculum state \\
\midrule
400 & 0.6331 & Selected $B_2$ checkpoint \\
450 & 0.6270 & $B_3$ \\
500 & 0.6171 & $B_3$ \\
550 & 0.6399 & Selected $B_3$ checkpoint \\
600 & 0.6444 & $B_4$ \\
650 & 0.6505 & $B_4$ \\
700 & 0.6558 & Selected $B_4$ checkpoint \\
750 & 0.6391 & $B_5$ + 50 updates \\
800 & 0.6353 & $B_5$ + 100 updates \\
\bottomrule
\end{tabular}
\end{table}

The complete test trajectories for the fixed-medium curriculum are reported in Table~\ref{tab:fixed_medium_trajectories}.
In all three runs, the active difficulty band is fixed at $B_3=[0.375,0.625]$, while samples
are reselected online according to the evolving policy. Bold values denote the best
observed test accuracy along each trajectory. These best observed values are
retrospective maxima along the completed trajectories; test accuracy is not used to
guide training, early stopping, parameter updates, or sample selection.

The best observed fixed-medium test accuracies are 0.6459 at 450 effective updates
for Seed~1, 0.6422 at 700 effective updates for Seed~2, and 0.6467 at 650 effective
updates for Seed~3. These values correspond to the fixed-medium results summarized
in the multi-seed comparison in the main text.

\begin{table}[htbp]
\centering
\caption{Test trajectories under the fixed-medium curriculum for the three random
seeds.}
\label{tab:fixed_medium_trajectories}
\small
\begin{tabular}{cccc}
\toprule
Effective updates & Seed 1 & Seed 2 & Seed 3 \\
\midrule
0    & 0.6141 & 0.6141 & 0.6141 \\
50   & 0.6181 & 0.6179 & 0.6133 \\
100  & 0.6195 & 0.6194 & 0.6118 \\
150  & 0.6223 & 0.6209 & 0.6171 \\
200  & 0.6274 & 0.6232 & 0.6224 \\
250  & 0.6344 & 0.6270 & 0.6103 \\
300  & 0.6406 & 0.6308 & 0.6368 \\
350  & 0.6384 & 0.6384 & 0.6384 \\
400  & 0.6376 & 0.6338 & 0.6096 \\
450  & \textbf{0.6459} & 0.6224 & 0.6171 \\
500  & 0.6452 & 0.6376 & 0.6247 \\
550  & 0.6217 & 0.6331 & 0.6293 \\
600  & 0.6217 & 0.6300 & 0.6399 \\
650  & 0.6247 & 0.6308 & \textbf{0.6467} \\
700  & -- & \textbf{0.6422} & 0.6437 \\
750  & -- & 0.6399 & 0.6422 \\
800  & -- & 0.6414 & 0.6141 \\
850  & -- & 0.6399 & 0.6141 \\
900  & -- & 0.6338 & 0.6331 \\
950  & -- & -- & 0.6262 \\
1000 & -- & -- & 0.6328 \\
\bottomrule
\end{tabular}
\end{table}

\FloatBarrier
\end{appendices}

\bibliography{references}

@article{shao2024deepseekmath,
  title        = {{DeepSeekMath}: Pushing the Limits of Mathematical Reasoning in Open Language Models},
  author       = {Shao, Zhihong and Wang, Peiyi and Zhu, Qihao and Xu, Runxin and Song, Junxiao and Bi, Xiao and Zhang, Haowei and Zhang, Mingchuan and Li, Y. K. and Wu, Y. and Guo, Daya},
  journal      = {arXiv preprint},
  year         = {2024},
  eprint       = {2402.03300},
  archivePrefix = {arXiv},
  url          = {https://arxiv.org/abs/2402.03300}
}

@inproceedings{bengio2009curriculum,
  title        = {Curriculum Learning},
  author       = {Bengio, Yoshua and Louradour, J{\'e}r{\^o}me and Collobert, Ronan and Weston, Jason},
  booktitle    = {Proceedings of the 26th Annual International Conference on Machine Learning},
  pages        = {41--48},
  year         = {2009},
  publisher    = {Association for Computing Machinery},
  doi          = {10.1145/1553374.1553380}
}

@article{cobbe2021gsm8k,
  title        = {Training Verifiers to Solve Math Word Problems},
  author       = {Cobbe, Karl and Kosaraju, Vineet and Bavarian, Mohammad and Chen, Mark and Jun, Heewoo and Kaiser, Lukasz and Plappert, Matthias and Tworek, Jerry and Hilton, Jacob and Nakano, Reiichiro and Hesse, Christopher and Schulman, John},
  journal      = {arXiv preprint},
  year         = {2021},
  eprint       = {2110.14168},
  archivePrefix = {arXiv},
  url          = {https://arxiv.org/abs/2110.14168}
}

@article{guo2025deepseekr1,
  title        = {{DeepSeek-R1} Incentivizes Reasoning in {LLMs} Through Reinforcement Learning},
  author       = {Guo, Daya and Yang, Dejian and Zhang, Haowei and Song, Junxiao and Wang, Peiyi and Zhu, Qihao and Xu, Runxin and Zhang, Ruoyu and Ma, Shirong and Bi, Xiao and others},
  journal      = {Nature},
  volume       = {645},
  pages        = {633--638},
  year         = {2025},
  doi          = {10.1038/s41586-025-09422-z},
  url          = {https://doi.org/10.1038/s41586-025-09422-z}
}

@inproceedings{xin2025deepseekprover,
  title        = {{DeepSeek-Prover-V1.5}: Harnessing Proof Assistant Feedback for Reinforcement Learning and Monte-Carlo Tree Search},
  author       = {Xin, Huajian and Ren, Z. Z. and Song, Junxiao and Shao, Zhihong and Zhao, Wanjia and Wang, Haocheng and Liu, Bo and Zhang, Liyue and Lu, Xuan and Du, Qiushi and Gao, Wenjun and Zhang, Haowei and Zhu, Qihao and Yang, Dejian and Gou, Zhibin and Wu, Z. F. and Luo, Fuli and Ruan, Chong},
  booktitle    = {International Conference on Learning Representations},
  year         = {2025},
  url          = {https://arxiv.org/abs/2408.08152}
}

@article{limr,
  title        = {{LIMR}: Less is More for {RL} Scaling},
  author       = {Li, Xuefeng and Zou, Haoyang and Liu, Pengfei},
  journal      = {arXiv preprint},
  year         = {2025},
  eprint       = {2502.11886},
  archivePrefix = {arXiv},
  url          = {https://arxiv.org/abs/2502.11886}
}

@article{adarft,
  title        = {Efficient Reinforcement Finetuning via Adaptive Curriculum Learning},
  author       = {Shi, Taiwei and Wu, Yiyang and Song, Linxin and Zhou, Tianyi and Zhao, Jieyu},
  journal      = {Transactions on Machine Learning Research},
  year         = {2026},
  eprint       = {2504.05520},
  archivePrefix = {arXiv},
  url          = {https://arxiv.org/abs/2504.05520}
}

@article{adacurl,
  title        = {{AdaCuRL}: Adaptive Curriculum Reinforcement Learning with Invalid Sample Mitigation and Historical Revisiting},
  author       = {Li, Renda and Huang, Hailang and Wei, Fei and Xiong, Feng and Wang, Yong and Chu, Xiangxiang},
  journal      = {Proceedings of the AAAI Conference on Artificial Intelligence},
  volume       = {40},
  number       = {27},
  pages        = {23123--23131},
  year         = {2026},
  doi          = {10.1609/aaai.v40i27.39479},
  url          = {https://ojs.aaai.org/index.php/AAAI/article/view/39479}
}

@inproceedings{e2hreasoner,
  title        = {Curriculum Reinforcement Learning from Easy to Hard Tasks Improves {LLM} Reasoning},
  author       = {Parashar, Shubham and Gui, Shurui and Li, Xiner and Ling, Hongyi and Vemuri, Sushil and Olson, Blake and Li, Eric and Zhang, Yu and Caverlee, James and Kalathil, Dileep and Ji, Shuiwang},
  booktitle    = {International Conference on Learning Representations},
  year         = {2026},
  url          = {https://arxiv.org/abs/2506.06632}
}

@inproceedings{dots,
  title        = {Improving Data Efficiency for {LLM} Reinforcement Fine-Tuning Through Difficulty-Targeted Online Data Selection and Rollout Replay},
  author       = {Sun, Yifan and Shen, Jingyan and Wang, Yibin and Chen, Tianyu and Wang, Zhendong and Zhou, Mingyuan and Zhang, Huan},
  booktitle    = {Advances in Neural Information Processing Systems},
  volume       = {38},
  year         = {2025},
  url          = {https://papers.nips.cc/paper_files/paper/2025/hash/fd54731cc635e2ed6033e8a0e193eddf-Abstract-Conference.html}
}

@inproceedings{depo,
  title        = {Towards High Data Efficiency in Reinforcement Learning with Verifiable Reward},
  author       = {Tang, Xinyu and Zhang, Zhenduo and Liu, Yurou and Zhao, Wayne Xin and Wen, Zujie and Zhang, Zhiqiang and Zhou, Jun},
  booktitle    = {International Conference on Learning Representations},
  year         = {2026},
  url          = {https://arxiv.org/abs/2509.01321}
}

@inproceedings{dps,
  title        = {Dynamics-Predictive Sampling for Active {RL} Finetuning of Large Reasoning Models},
  author       = {Mao, Yixiu and Qu, Yun and Wang, Qi and Zou, Heming and Ji, Xiangyang},
  booktitle    = {International Conference on Learning Representations},
  year         = {2026},
  url          = {https://arxiv.org/abs/2603.10887}
}

@inproceedings{bots,
  title        = {{BOTS}: A Unified Framework for Bayesian Online Task Selection in {LLM} Reinforcement Finetuning},
  author       = {Shen, Qianli and Chen, Daoyuan and Huang, Yilun and Ling, Zhenqing and Li, Yaliang and Ding, Bolin and Zhou, Jingren},
  booktitle    = {International Conference on Learning Representations},
  year         = {2026},
  url          = {https://arxiv.org/abs/2510.26374}
}

@article{insight,
  title        = {Efficient {RLVR} Training via Weighted Mutual Information Data Selection},
  author       = {Zhou, Xinyu and Zhu, Boyu and Zhang, Haotian and Wang, Huiming and Guo, Zhijiang},
  journal      = {arXiv preprint},
  year         = {2026},
  eprint       = {2603.01907},
  archivePrefix = {arXiv},
  url          = {https://arxiv.org/abs/2603.01907}
}

@inproceedings{odf,
  title        = {Online Difficulty Filtering for Reasoning Oriented Reinforcement Learning},
  author       = {Bae, Sanghwan and Hong, Jiwoo and Lee, Min Young and Kim, Hanbyul and Nam, Jeongyeon and Kwak, Donghyun},
  booktitle    = {Proceedings of the 19th Conference of the European Chapter of the Association for Computational Linguistics (Volume 1: Long Papers)},
  pages        = {700--719},
  year         = {2026},
  address      = {Rabat, Morocco},
  publisher    = {Association for Computational Linguistics},
  doi          = {10.18653/v1/2026.eacl-long.30},
  url          = {https://aclanthology.org/2026.eacl-long.30/}
}

@inproceedings{pcl,
  title        = {Prompt Curriculum Learning for Efficient {LLM} Post-Training},
  author       = {Gao, Zhaolin and Kim, Joongwon and Sun, Wen and Joachims, Thorsten and Wang, Sid and Pang, Richard Yuanzhe and Tan, Liang},
  booktitle    = {International Conference on Learning Representations},
  year         = {2026},
  url          = {https://arxiv.org/abs/2510.01135}
}

@inproceedings{cures,
  title        = {{CurES}: From Gradient Analysis to Efficient Curriculum Learning for Reasoning {LLMs}},
  author       = {Zeng, Yongcheng and Sun, Zexu and Ji, Bokai and Min, Erxue and Cai, Hengyi and Wang, Shuaiqiang and Yin, Dawei and Zhang, Haifeng and Chen, Xu and Wang, Jun},
  booktitle    = {International Conference on Learning Representations},
  year         = {2026},
  url          = {https://arxiv.org/abs/2510.01037}
}

@inproceedings{reschedule,
  title        = {Scheduling Your {LLM} Reinforcement Learning with Reasoning Trees},
  author       = {Wang, Hong and Hao, Zhezheng and Luo, Jian and Wei, Chenxing and Shu, Yao and Liu, Lei and Lin, Qiang and Dong, Hande and Chen, Jiawei},
  booktitle    = {International Conference on Learning Representations},
  year         = {2026},
  url          = {https://arxiv.org/abs/2510.24832}
}

@inproceedings{mopps,
  title        = {Can Prompt Difficulty be Online Predicted for Accelerating {RL} Finetuning of Reasoning Models?},
  author       = {Qu, Yun and Wang, Qi and Mao, Yixiu and Hu, Vincent Tao and Ommer, Bj{\"o}rn and Ji, Xiangyang},
  booktitle    = {Proceedings of the 32nd {ACM} {SIGKDD} Conference on Knowledge Discovery and Data Mining},
  year         = {2026},
  pages        = {1240--1250},
  doi          = {10.1145/3770854.3780263},
  eprint       = {2507.04632},
  archivePrefix = {arXiv},
  url          = {https://arxiv.org/abs/2507.04632}
}

@inproceedings{uforl,
  title        = {{UFO-RL}: Uncertainty-Focused Optimization for Efficient Reinforcement Learning Data Selection},
  author       = {Zhao, Yang and Xiong, Kai and Ding, Xiao and Du, Li and Ouyang, Yangou and Sun, Zhouhao and Guan, Jiannan and Zhang, Wenbin and Liu, Bin and Hu, Dong and Qin, Bing and Liu, Ting},
  booktitle    = {Advances in Neural Information Processing Systems},
  volume       = {38},
  year         = {2025},
  url          = {https://arxiv.org/abs/2505.12457}
}

@inproceedings{rftinf,
  title        = {Understanding Data Influence in Reinforcement Finetuning},
  author       = {Tan, Haoru and Wu, Xiuzhe and Wu, Sitong and Zhang, Shaofeng and Chen, Yanfeng and Sun, Xingwu and Shen, Jeanne and Qi, Xiaojuan},
  booktitle    = {Advances in Neural Information Processing Systems},
  volume       = {38},
  year         = {2025},
  doi          = {10.52202/085713-1294},
  url          = {https://proceedings.neurips.cc/paper_files/paper/2025/hash/3755a02b1035fbadd5f93a022170e46f-Abstract-Conference.html}
}

@article{qwen25technical,
  title        = {{Qwen2.5} Technical Report},
  author       = {Yang, An and Yang, Baosong and Zhang, Beichen and Hui, Binyuan and Zheng, Bo and Yu, Bowen and Li, Chengyuan and Liu, Dayiheng and Huang, Fei and Wei, Haoran and others},
  journal      = {arXiv preprint},
  year         = {2024},
  eprint       = {2412.15115},
  archivePrefix = {arXiv},
  url          = {https://arxiv.org/abs/2412.15115}
}

@inproceedings{hu2021lora,
  title        = {{LoRA}: Low-Rank Adaptation of Large Language Models},
  author       = {Hu, Edward J. and Shen, Yelong and Wallis, Phillip and Allen-Zhu, Zeyuan and Li, Yuanzhi and Wang, Shean and Wang, Lu and Chen, Weizhu},
  booktitle    = {International Conference on Learning Representations},
  year         = {2022},
  url          = {https://arxiv.org/abs/2106.09685}
}

@inproceedings{xu-etal-2026-prune,
    title = {Prune as You Generate: Online Rollout Pruning for Faster and Better {RLVR}},
    author = {Xu, Haobo and Chen, Sirui and Qiu, Ruizhong and Yan, Yuchen and
              Luo, Chen and Cheng, Monica Xiao and He, Jingrui and Tong, Hanghang},
    booktitle = {Proceedings of the 64th Annual Meeting of the Association for Computational Linguistics (Volume 1: Long Papers)},
    year = {2026},
    pages = {13876--13893},
    publisher = {Association for Computational Linguistics},
    doi = {10.18653/v1/2026.acl-long.632}
}

@inproceedings{le2025noprompt,
    title = {No Prompt Left Behind: Exploiting Zero-Variance Prompts in {LLM} Reinforcement Learning via Entropy-Guided Advantage Shaping},
    author = {Le, Thanh-Long V. and Jeon, Myeongho and Vu, Kim and Lai, Viet and Yang, Eunho},
    booktitle = {International Conference on Learning Representations},
    year = {2026},
    eprint = {2509.21880},
    archivePrefix = {arXiv},
    url = {https://arxiv.org/abs/2509.21880}
}

@inproceedings{dapo,
  title={{DAPO}: An Open-Source {LLM} Reinforcement Learning System at Scale},
  author={Yu, Qiying and Zhang, Zheng and Zhu, Ruofei and Yue, Yufeng and
  Chen, Wenhu and others},
  booktitle={Advances in Neural Information Processing Systems},
  year={2025},
  doi={10.52202/085713-3775},
  eprint={2503.14476},
  archivePrefix={arXiv},
  url={https://arxiv.org/abs/2503.14476}
}

@article{drgrpo,
  title={Understanding {R1-Zero-Like} Training: A Critical Perspective},
  author={Liu, Zichen and Chen, Changyu and Li, Wenjun and others},
  journal={arXiv preprint},
  year={2025},
  eprint={2503.20783},
  archivePrefix={arXiv},
  url={https://arxiv.org/abs/2503.20783}
}

@article{gspo,
  title={Group Sequence Policy Optimization},
  author={Zheng, Chujie and Liu, Shixuan and Li, Mingze and others},
  journal={arXiv preprint},
  year={2025},
  eprint={2507.18071},
  archivePrefix={arXiv},
  url={https://arxiv.org/abs/2507.18071}
}

\end{document}